\documentclass{article}
\usepackage{iclr2027_conference}
\iclrfinalcopy

\usepackage[T1]{fontenc}
\usepackage{times}
\usepackage{amsmath,amssymb,graphicx,booktabs,array,multirow}
\usepackage{algorithm,algpseudocode}
\usepackage{tikz}
\usetikzlibrary{arrows.meta,positioning,fit,calc}
\definecolor{intentblue}{HTML}{2673A6}
\definecolor{execorange}{HTML}{BD6D28}
\definecolor{anchorgray}{HTML}{52616D}
\definecolor{lightblue}{HTML}{EDF5FA}
\definecolor{lightorange}{HTML}{FCF1E6}
\usepackage[breaklinks,colorlinks,allcolors=intentblue]{hyperref}
\newcommand{\R}{\mathbb{R}}

\newcommand{\method}{LeapBot-JAC}
\tikzset{arr/.style={-{Stealth[length=1.7mm]},semithick,draw=anchorgray},box/.style={draw=anchorgray,rounded corners=2pt,align=center,inner sep=5pt,font=\sffamily\small},rb/.style={box,draw=intentblue,fill=lightblue},ab/.style={box,draw=execorange,fill=lightorange}}

\usepackage{pgfplots}
\pgfplotsset{compat=1.18}

\title{\centering LeapBot-JAC: Joint Action Chains for World Action Models}

\author{
\makebox[\textwidth][c]{
\begin{tabular}{c}
Chenhao Zhu$^{1}$\thanks{Equal contribution.}, Baibing Liu$^{1}$\footnotemark[1], Pei Liu
$^{2}$, Rui Yang$^{2}$, Lang Zhang$^{1\dagger}$\thanks{Project leader.}\thanks{Corresponding author.}, \\Mingyue Feng$^{1}$, Yaonong Wang$^{1}$, Jun Ma$^{2,3}$\\
$^1$Leapmotor \\
$^2$The Hong Kong University of Science and Technology (Guangzhou) \\
$^3$The Hong Kong University of Science and Technology \\
\texttt{zhang\_lang7@leapmotor.com} \\
Code: \url{https://github.com/LeapWM/leapbot-jac}.
\end{tabular}
}
}

\begin{document}
\maketitle
\begin{abstract}
Action-chunking policies predict future controls but execute only a prefix before replanning. How should a fixed action-token budget be organized to support execution? We propose LeapBot-JAC, which allocates sparse future-control references and dense execution targets within one shared action expert. Both chains are supervised by demonstrated controls and jointly denoised at aligned physical-time positions. Directed attention allows execution tokens to read reference tokens subject to the host's visibility constraints. JAC adds neither action tokens nor parameters and retains the host's world-modeling process; both chains participate in inference, while only the execution chain drives the robot. On LIBERO, JAC reaches 98.3\% success with FastWAM and LeapBot-WA, up from 97.0\% and 97.3\%. On zero-shot LIBERO-Plus, their success rates improve from 53.5\% to 70.4\% and from 73.1\% to 77.3\%, respectively. RoboTwin and two real-world tasks provide additional deployment evaluations. Single-training-seed comparisons and FastWAM ablations support action-role organization as a practical design choice under a fixed action-token budget. 
\end{abstract}

\section{Introduction}
Action-chunking policies predict a sequence of future robot controls, but closed-loop deployment often executes only a short prefix before querying the policy again. ACT and Diffusion Policy exploit sequence generation for temporally coherent behavior while using temporal ensembling or receding-horizon execution to incorporate new observations~\citep{zhao2023act,chi2023diffusion}. Predictions beyond the executed prefix can still influence the current output through the model's internal dependencies. The resulting design question is how to organize those predictions: should every token have the same temporal density and deployment role?

\begin{figure}[b]
\centering
\input{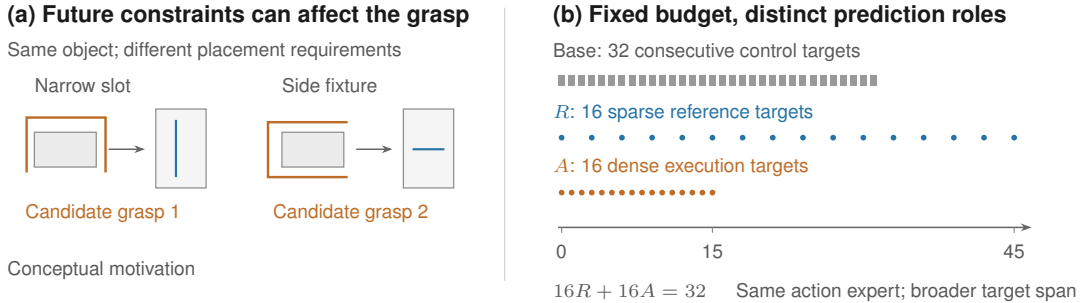}
\resizebox{\linewidth}{!}{%
\begin{tikzpicture}[x=1cm,y=1cm,every node/.style={jac text}]
\path[use as bounding box] (0,-.32) rectangle (13.65,3.68);
\node[jac title,anchor=west] at (0,3.44) {(a) Future constraints can affect the grasp};
\node[jac title,anchor=west] at (7.05,3.44) {(b) Fixed budget, distinct prediction roles};
\draw[jac line,draw=black!18] (6.55,.02)--(6.55,3.55);
\node[jac small,anchor=west] at (0,2.99) {Same object; different placement requirements};
\foreach \x/\txt in {.35/Narrow slot,3.55/Side fixture}{
 \node[jac small,anchor=west] at (\x,2.55) {\txt};
 \draw[jac line,fill=black!8] (\x+.12,1.49) rectangle ++(.8,.45);
 \draw[jac arrow] (\x+1.08,1.72)--(\x+1.55,1.72);
}
\draw[draw=execorange,line width=.9pt] (.37,1.42)--(.37,2.13)--(1.39,2.13)--(1.39,1.42);
\draw[draw=execorange,line width=.9pt] (4.52,2.07)--(3.49,2.07)--(3.49,1.36)--(4.52,1.36);
\draw[jac line,fill=black!4] (2.04,1.22) rectangle (2.69,2.24);
\draw[intentblue,line width=.8pt] (2.3,1.36)--(2.3,2.1);
\draw[jac line,fill=black!4] (5.24,1.22) rectangle (5.89,2.24);
\draw[intentblue,line width=.8pt] (5.36,1.72)--(5.77,1.72);
\node[jac small,text=execorange] at (1.36,.89) {Candidate grasp 1};
\node[jac small,text=execorange] at (4.56,.89) {Candidate grasp 2};
\node[jac small,anchor=west] at (0,.16) {Conceptual motivation};

\node[jac small,anchor=west] at (7.05,2.97) {Base: 32 consecutive control targets};
\foreach \j in {0,...,31}{\fill[black!40] (7.24+.13*\j,2.54) rectangle ++(.09,.13);}
\node[jac small,text=intentblue,anchor=west] at (7.05,2.19) {$R$: 16 sparse reference targets};
\foreach \j in {0,...,15}{\fill[intentblue] (7.285+.39*\j,1.87) circle (.037);}
\node[jac small,text=execorange,anchor=west] at (7.05,1.49) {$A$: 16 dense execution targets};
\foreach \j in {0,...,15}{\fill[execorange] (7.285+.13*\j,1.16) circle (.037);}
\draw[jac arrow] (7.24,.71)--(13.36,.71);
\foreach \j in {0,15,45}{
 \draw[jac line] (7.285+.13*\j,.71)--++(0,-.07);
 \node[jac small,below] at (7.285+.13*\j,.61) {\j};
}
\node[jac small,anchor=west] at (7.05,-.13) {$16R+16A=32$\quad\; Same action expert; broader target span};
\end{tikzpicture}}
\caption{\textbf{Motivation and budget allocation.} (a) A conceptual illustration of how future placement constraints may affect current grasp choices; it is not an experimental rollout or a validated feasible solution. (b) JAC reallocates 32 action tokens into sparse reference and dense execution targets on the same control-time axis. Only the execution prefix drives the robot.}
\label{fig:motivation}
\end{figure}

World action models (WAMs) couple robot control with learning environment dynamics. Motus supports multiple video--action modeling modes~\citep{bi2025motus}; LingBot-VA uses causal video--action generation~\citep{li2026lingbotva}; Fast-WAM retains video co-training while omitting future-video generation at inference~\citep{yuan2026fastwam}. These approaches primarily organize world representations and their interaction with control. Within an existing action expert, the organization of a fixed number of control predictions remains a separate design choice.

We study \textbf{action-role organization under a fixed action-token budget}. Dense predictions preserve near-term control resolution, whereas sparsely sampled control targets span a longer interval using fewer tokens. Combining these roles provides a way to expose execution to a temporally extended reference without adding an action expert or expanding its token sequence. The reference consists of robot controls, not semantic subgoals or executable waypoints; its role is to condition the jointly generated execution chain.

We propose \method{} (Joint Action Chains, JAC). A sparse reference chain $R$ and a dense execution chain $A$ share input projections, Transformer layers, an output head, and every action forward pass. Both chains are supervised by demonstrated controls and evolve together during denoising, with independent noise samples at a common noise level. Physical-time positions align the two target sequences. A directed role mask permits $A$ to read $R$ while blocking direct access in the reverse direction, subject to the host's legal attention edges. Only the execution prefix is sent to the robot. This construction reorganizes action prediction while preserving the host's world-modeling process.

Prior work already explores joint generation of high-level intent and executable actions~\citep{li2026lawa,hao2025chd}. JAC's focus is more specific: directly supervised control references and dense execution coexist within one existing action expert at a fixed token budget. We assess task success across two hosts and examine how information-flow direction, reference weight, and reference stride affect performance.

Our contributions are threefold. \textbf{(1) Action-role allocation:} we formulate sparse reference and dense execution prediction within a fixed action-token budget, separating the supervision span from the executed prefix. \textbf{(2) Shared-expert joint generation:} we combine control-label supervision, physical-time alignment, and directed cross-chain access without adding action parameters, while retaining host visibility constraints. \textbf{(3) Empirical evidence on two hosts:} JAC improves LIBERO success for FastWAM and LeapBot-WA to 98.3\%, and improves their LIBERO-Plus totals by 16.9 and 4.2 percentage points, respectively. Three FastWAM ablations characterize sensitivity to access direction, reference supervision, and temporal coverage.

\section{Related Work}
\paragraph{Action sequences and hierarchical control.}
ACT improves fine manipulation through sequence generation and temporal ensembling~\citep{zhao2023act}. Diffusion Policy represents multimodal action distributions with conditional diffusion and receding-horizon control~\citep{chi2023diffusion}. Play-LMP learns latent plans from play data for goal-conditioned behavior~\citep{lynch2019play}. Dense Policy progressively expands sparse action keyframes into a dense sequence~\citep{su2025densepolicy}. JAC instead retains simultaneous reference and execution roles at a fixed total token count. Hierarchical methods distinguish longer-term plans from local execution. CHD jointly diffuses high-level subgoals and low-level trajectories, allowing low-level feedback to refine subgoals~\citep{hao2025chd}. JAC allocates reference and execution roles inside one existing action expert; its directionality is evaluated as an architectural choice.

\paragraph{World modeling and action generation.}
Motus combines multiple experts and modality schedules for different conditional generation modes~\citep{bi2025motus}. LingBot-VA jointly models video and control with causal blocks and history updates~\citep{li2026lingbotva}. Fast-WAM separates video co-training from test-time future generation~\citep{yuan2026fastwam}, while LeapBot-WA explores world anchoring in predictive semantic space~\citep{liu2026leapbotwa}. JAC preserves a host's world objective and sampling process while changing targets and visibility within its action budget. Beyond joint action generators, DINO-WM optimizes actions through predicted visual features~\citep{zhou2024dinowm}, and V-JEPA 2-AC uses an action-conditioned latent world model for planning~\citep{assran2025vjepa2}.

\paragraph{Intermediate representations and action reasoning.}
UniPi synthesizes text-conditioned video plans and extracts controls from them~\citep{du2023unipi}; CoTDiffusion generates visual subgoals to guide a low-level policy~\citep{ni2024cotdiffusion}. These differ from JAC's directly supervised control references. ECoT generates embodied reasoning about objects, end-effector states, and subtasks before actions~\citep{zawalski2024ecot}. LAWA learns codebook-based transition targets from action-free video, jointly denoises separate latent-action and executable-action experts, and omits the future video branch at inference~\citep{li2026lawa}. Its online latent intent evolves in a continuous embedding space. JAC shares the idea that execution can use evolving intent, but differs in \textbf{supervision}, taken directly from robot controls without a new video-action tokenizer; \textbf{parameters and budget}, reallocating a fixed token count inside an existing expert; and \textbf{host integration}, retaining the host world process while changing action-target organization and direct cross-chain access.

\paragraph{Spatial, semantic, and historical conditions.}
Robust-WAM aligns action-stream queries with future semantic features and their physical times~\citep{yan2026robustwam}. MaskWAM combines object-mask prompting and prediction~\citep{yu2026maskwam}, and MemoryWAM studies persistent history~\citep{yang2026memorywam}. These mechanisms address semantic, spatial, and historical conditioning. JAC complements these mechanisms by supervising intermediate robot controls.

\paragraph{Noise schedules and sequence generation.}
Diffusion Forcing assigns independent noise levels to sequence tokens~\citep{chen2024diffusionforcing}. JAC instead uses independent noise realizations at a common action noise level.

\section{Joint Action Chains under a Fixed Budget}
\subsection{Problem and Fixed-Budget Formulation}
At query time $t$, conditions $c_t$ comprise observations, instructions, proprioception, and host-permitted history. Let $U=[u_0,\ldots,u_{T-1}]\in\R^{T\times d}$ denote demonstrated future controls relative to $t$. We distinguish supervision-window length $T$, action-token budget $B$, and executed prefix $K$. JAC changes target allocation, physical-time positions, cross-chain visibility, and reference supervision within one action expert.

Let $W,C$ denote host world states and caches, and $M_{\rm host}$ its legal visibility. JAC retains the host world process and requires configurable action targets, physical-time positions, and attention visibility. Reference times must fit within a legally joint generation block; host integration is detailed in Appendix~\ref{app:integration}.

\begin{figure*}[t]
\centering
\includegraphics[width=1.\textwidth]{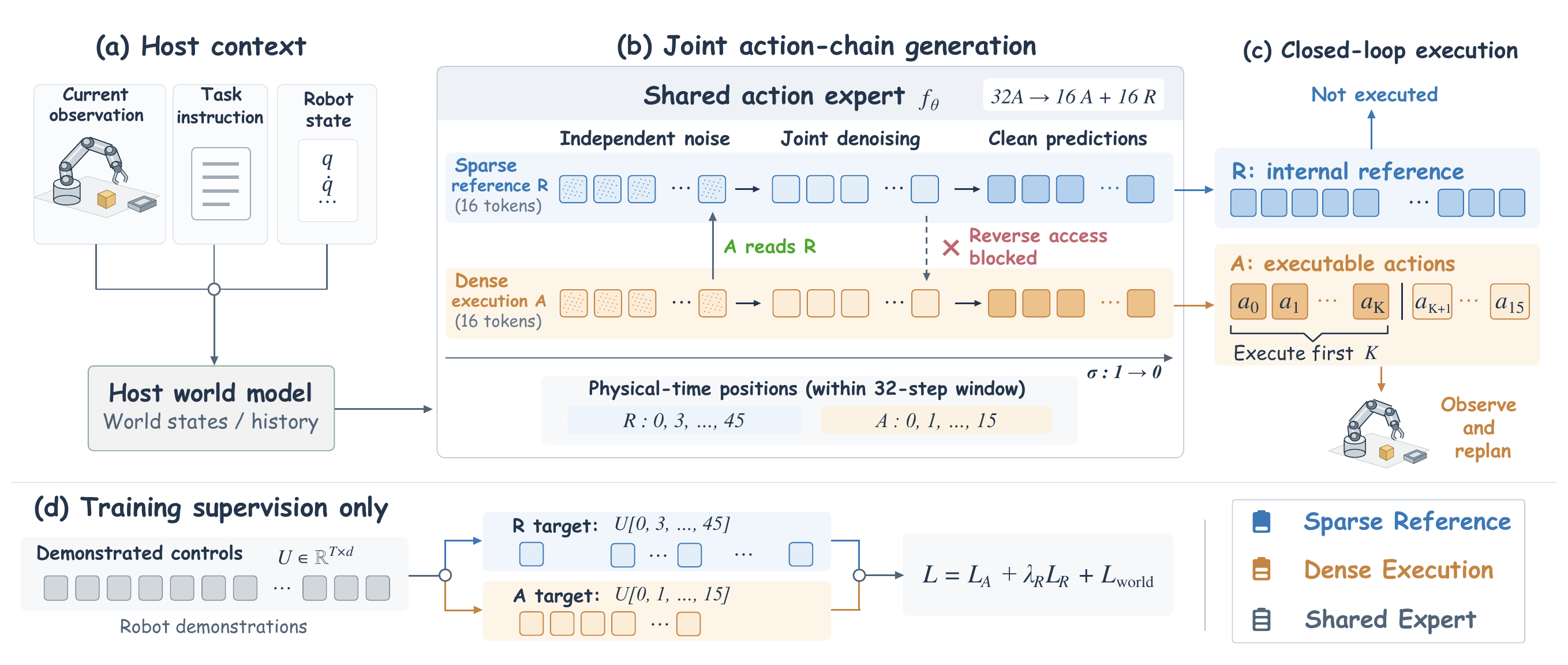}
\caption{Overview of LeapBot-JAC. A shared action expert jointly denoises a sparse reference chain \(R\) and a dense execution chain \(A\) within a fixed action-token budget. Physical-time encoding aligns the two chains, while a directed attention mask allows \(A\) to attend to \(R\) and blocks direct reverse access, subject to the host's visibility constraints. Both chains are supervised by demonstrated controls, and only the execution prefix drives the robot. The illustrated $K=10$ is used on LIBERO/LIBERO-Plus; RoboTwin uses $K=16$. The host's world-modeling objectives and sampling schedule are preserved.}
\label{fig:architecture}
\end{figure*}

\subsection{Target Construction and Boundary Padding}
A fixed rule $\pi$ produces ordered index sets $I_R,I_A$, with $|I_R|=n_R$, $|I_A|=n_A$, and $n_R+n_A=B$. Target allocation is
\begin{equation}
G_\pi(U)=\big(U[I_R],U[I_A],I_R,I_A\big).
\label{eq:allocation}
\end{equation}
Execution uses $I_A=\{0,\ldots,n_A-1\}$, with $K\leq n_A$, while $I_R=\{0,s,\ldots,(n_R-1)s\}$ sparsely samples future controls. Unless an ablation specifies otherwise, we use $s=3$, $\lambda_R=0.2$, $B=32$, $n_R=n_A=16$, and $K=10$ on LIBERO/LIBERO-Plus. RoboTwin uses $K=16$ for both bases and JAC. The default loader window has $T=48$ controls; its 16 sampled references end at index 45. The action-token budget remains 32:
\begin{equation}
R_0=[u_0,u_3,\ldots,u_{45}],\quad A_0=[u_0,u_1,\ldots,u_{15}].
\label{eq:targets}
\end{equation}
Subscript 0 denotes a clean target in normalized action space; indices denote physical control times. Both chains are subsequences of one expert. Across base/JAC and $s=2,3,4$, trajectory starts, video-frame indices and counts, world-prediction steps, and world-loss coefficients remain fixed. Extending reference coverage changes neither world targets nor the executed prefix.

\paragraph{Boundary padding.}
Out-of-episode indices are clamped to the last valid action and marked by $P$, with $P_R=P[:,::3]$ and $P_A=P[:,:16]$ at the default stride. Filled targets undergo normal transformations and noising. Padding is excluded from direct supervision but remains attention-visible under the role mask. Appendix~\ref{app:padding} specifies the delta-control fill rule.

\paragraph{Coverage and overlap.}
For a fully valid window, $R$ and $A$ overlap at $\{0,3,6,9,12,15\}$: 32 tokens cover 26 distinct control times. References supervise both overlapping and more distant sampled times; padding reduces valid coverage. In general,
\begin{equation}
O=|I_R\cap I_A|,\qquad D=|I_R\cup I_A|=B-O.
\end{equation}
Overlap and distinct-target count are coupled at fixed budget. Dense $A$ preserves near-term control resolution; sparse $R$ supplies auxiliary targets over a longer physical span. These incremental controls are predictive references, not executable waypoints or abstract subgoals.

\subsection{Shared Generation and Time Alignment}
We retain the host's continuous generation parameterization. For linear flow matching~\citep{lipman2023flow}, each sample uses a common noise level $\sigma\in[0,1]$ and independent Gaussian samples $\epsilon_R,\epsilon_A$:
\begin{align}
R_\sigma&=(1-\sigma)R_0+\sigma\epsilon_R,\\
A_\sigma&=(1-\sigma)A_0+\sigma\epsilon_A.
\end{align}
The existing action expert receives $X_\sigma=[R_\sigma;A_\sigma]$:
\begin{equation}
[\widehat v_R;\widehat v_A]=f_\theta(X_\sigma,\sigma,p;W,C,c_t,M).
\label{eq:forward}
\end{equation}
Both chains share input/output projections, time embeddings, Transformer layers, and each action forward pass. Their targets, positions, noise, losses, and visibility distinguish their roles. Independent noise differentiates noisy inputs at overlapping times; shared conditions and parameters still induce dependence.

\begin{figure}[t]
\centering
\input{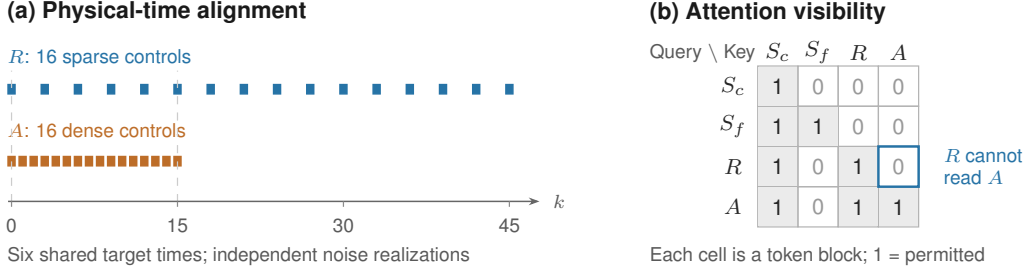}
\resizebox{\linewidth}{!}{%
\begin{tikzpicture}[x=1cm,y=1cm,every node/.style={jac text}]
\path[use as bounding box] (0,-.25) rectangle (13.65,3.42);
\node[jac title,anchor=west] at (0,3.17) {(a) Physical-time alignment};
\node[jac title,anchor=west] at (8.3,3.17) {(b) Attention visibility};
\node[jac small,text=intentblue,anchor=west] at (0,2.58) {$R$: 16 sparse controls};
\node[jac small,text=execorange,anchor=west] at (0,1.65) {$A$: 16 dense controls};
\foreach \j in {0,...,15}{
 \draw[jac r,fill=intentblue] (.13+.143*3*\j,2.12) rectangle ++(.09,.12);
 \draw[jac a,fill=execorange] (.13+.143*\j,1.19) rectangle ++(.09,.12);
}
\foreach \j in {0,15}{\draw[jac line,dashed,draw=black!20] (.175+.143*\j,.72)--(.175+.143*\j,2.42);}
\draw[jac arrow] (.13,.73)--(7,.73);
\foreach \j in {0,15,30,45}{
 \draw[jac line] (.175+.143*\j,.73)--++(0,-.07);
 \node[jac small,below] at (.175+.143*\j,.63) {\j};
}
\node[jac small,anchor=west] at (7.05,.73) {$k$};
\node[jac small,anchor=west] at (0,.02) {Six shared target times; independent noise realizations};

\node[jac small,anchor=west] at (8.3,2.65) {Query $\backslash$ Key};
\foreach \j/\lab in {0/S_c,1/S_f,2/R,3/A}{
 \node at (10.08+.52*\j,2.66) {$\lab$};
 \node at (9.5,2.22-.52*\j) {$\lab$};
}
\foreach \r in {0,...,3}{\foreach \c in {0,...,3}{
 \draw[jac line,draw=black!30] (9.82+.52*\c,2.48-.52*\r) rectangle ++(.52,-.52);
}}
\foreach \r/\c in {0/0,1/0,1/1,2/0,2/2,3/0,3/2,3/3}{
 \fill[black!8] (9.83+.52*\c,2.47-.52*\r) rectangle ++(.50,-.50);
 \node at (10.08+.52*\c,2.22-.52*\r) {1};
}
\foreach \r/\c in {0/1,0/2,0/3,1/2,1/3,2/1,2/3,3/1}{\node[text=black!40] at (10.08+.52*\c,2.22-.52*\r) {0};}
\draw[draw=intentblue,line width=.9pt] (11.38,1.44) rectangle (11.9,.92);
\node[jac small,align=left,anchor=west,text=intentblue] at (12.1,1.18) {$R$ cannot\\read $A$};
\node[jac small,anchor=west] at (8.3,.02) {Each cell is a token block; 1 = permitted};
\end{tikzpicture}}
\caption{\textbf{Time alignment and directed visibility.} (a) At $s=3$, reference indices span 0--45 and execution indices span 0--15. (b) The outlined $16\times16$ block prevents $R$ queries from reading $A$ keys. $S_c,S_f$ denote current/future semantics. The full matrix illustrates an action-independent semantic context; non-action restrictions are inherited from the host. General visibility follows Eq.~\eqref{eq:mask}.}
\label{fig:timemask}
\end{figure}

Figure~\ref{fig:timemask} places both chains on one physical-time axis. Their position vector is
\begin{equation}
p=[0,3,\ldots,45;\;0,1,\ldots,15].
\end{equation}
For hosts using RoPE~\citep{su2021roformer}, the relative offset between execution token $i$ and reference token $j$ is $3j-i$, rather than their concatenated-slot offset $j-(16+i)$. The host frequency table is retained, with no additional chain-identity embedding. Physical durations must be aligned when control frequencies differ.

\subsection{Directed Visibility and Training Objective}
Mapping host visibility to physical times and legal generation blocks gives
\begin{equation}
M_{\rm JAC}=M_{\rm host}\land M_{\rm role},\quad
M_{\rm role}|_{R,A}=\begin{bmatrix}1&0\\1&1\end{bmatrix}.
\label{eq:mask}
\end{equation}
Rows are queries and columns are keys: $A$ reads $R$, while $R$ cannot read $A$. For a fully visible 32-token action block, the mask adds one $16\times16$ zero block. Edges outside the action--action block retain host visibility.

Figure~\ref{fig:timemask} illustrates a host where semantic tokens do not read actions and actions cannot read future semantics. These additional restrictions are inherited from that host.

\paragraph{Local directionality.}
The mask blocks direct $A\rightarrow R$ attention. Indirect reverse paths through host states require the conditions in Appendix~\ref{app:dependency} to be excluded; directionality does not imply statistical independence.

\paragraph{Supervision strength.}
Let $P_{L,k}=1$ mark padding and $N_L=\sum_k(1-P_{L,k})$ count valid targets in chain $L$. The per-sample chain loss averages squared error over action dimensions and valid tokens, with the host's common timestep weight $w(\tau)$:
\begin{equation}
\ell_L=\frac{w(\tau)}{d\,\max(N_L,1)}
\sum_k(1-P_{L,k})\big\|\widehat v_{L,k}-(\epsilon_{L,k}-L_{0,k})\big\|_2^2,
\qquad L\in\{R,A\}.
\label{eq:chainloss}
\end{equation}
Here $\tau$ is the sampled training timestep, distinct from environment time $t$, and $\sigma$ is its scheduler-mapped noise level. Both chains share $\tau$ and $w(\tau)$; a chain with no valid targets contributes zero direct loss. The training objective is
\begin{equation}
\mathcal L=\mathbb E\!\left[\ell_A+\lambda_R\ell_R\right]+\mathcal L_{\rm host,world},
\label{eq:loss}
\end{equation}
where the expectation averages over samples, timesteps, and noise; world-loss coefficients are unchanged. With execution weight 1, $\lambda_R$ controls direct reference supervision, not attention strength. Overlapping times are supervised in both roles. Valid-token averaging removes direct scaling with target count, while padding can still alter coverage and predictions (Appendix~\ref{app:losscontrol}).

Execution losses also backpropagate through reference representations, including padded positions. Thus $\lambda_R=0$ would retain reference computation and indirect supervision, and is not equivalent to the base. 

\subsection{Inference and Resource Accounting}
Each query initializes world states and caches as in the host, and $R,A$ with independent Gaussian noise. Both chains are jointly integrated from $\sigma=1$ to 0 in normalized action space. Only $\widehat A$ is inverse-transformed, and its first $K$ controls are executed before replanning. Inference has no demonstration-tail padding mask: all reference and execution slots are generated at every query. Every reference slot participates in each action-expert update even when the reference trajectory is not returned.

Algorithm~\ref{alg:jac} in Appendix~\ref{app:sampling} summarizes the sampling procedure.

At fixed $B$, JAC preserves action parameter count, tokens per step, and action-expert forward counts while retaining host world updates. Structural equality does not by itself imply equal runtime; model-call latency and memory accounting are addressed in Section~\ref{sec:cost}.

\section{Experiments}
We evaluate whether JAC improves two WAM hosts, FastWAM and LeapBot-WA, across standard manipulation, zero-shot perturbations, and bimanual tasks. Controlled LIBERO ablations examine visibility, supervision, and reference stride; two real-world tasks test physical deployment.

\subsection{Experimental Setup}
\paragraph{Configuration and evaluation.}
Unless otherwise specified, JAC uses $16R+16A$, reference stride $s=3$, supervision weight $\lambda_R=0.2$, and directed $R\rightarrow A$ access. Our local Base/JAC comparisons fix the total action-token count at 32. LIBERO/LIBERO-Plus use $K=10$; RoboTwin bases and JAC are all re-evaluated with $K=16$, matching JAC's execution-chain length. LIBERO~\citep{liu2023libero} reports success on Spatial, Object, Goal, and Long, with their four-suite mean. For LIBERO-Plus~\citep{fei2026liberoplus}, we directly evaluate LIBERO-trained policies without Plus training, fine-tuning, or test-time parameter adaptation. Its seven perturbation categories measure robustness of learned tasks; our Total aggregates all test episodes. RoboTwin 2.0~\citep{chen2025robotwin2} reports Clean, Random, and their mean.

\paragraph{Baselines and reporting.}
We compare our reproduced FastWAM and LeapBot-WA bases with their JAC variants. External references are $\pi_0$, $\pi_{0.5}$, Motus, LingBot-VA, LAWA, and ImageWAM, using available results per benchmark. Table citations identify the numerical source, prioritizing original reports and using later evaluations when needed. Literature results provide context under different training recipes and model budgets; within-host pairs assess JAC. Parentheses denote absolute percentage-point changes from each reproduced base. We follow each host paper's training hyperparameters~\citep{yuan2026fastwam,liu2026leapbotwa}, with the JAC and execution settings specified here. Training and evaluation use an eight-NVIDIA-H100 system. Appendix~\ref{app:parameters} summarizes the protocol.

\subsection{Main Results}
\begin{table*}[!t]
\centering
\caption{\textbf{Standard manipulation.}
Success rates (\%) on LIBERO and RoboTwin 2.0.
Cited results and reproduced Base/JAC pairs are separated.
Parentheses denote the improvement in average success rate (pp);
bold indicates the better result within each reproduced pair.
Our RoboTwin experiments execute 16 actions from each predicted action chunk;
literature protocols are retained as reported. Motus LIBERO scores are reported by \citet{yuan2026fastwam}; its RoboTwin scores follow \citet{bi2025motus}.}
\label{tab:main}

\vspace{2pt}

\fontsize{8.3}{10}\selectfont
\setlength{\tabcolsep}{4.5pt}
\renewcommand{\arraystretch}{1.08}

\newcommand{\avgval}[1]{%
  \makebox[5.8em][c]{#1}%
}
\newcommand{\jacavg}[2]{%
  \makebox[5.8em][c]{%
    \textbf{#1}\,
    {\color{anchorgray}\fontsize{6.5}{7}\selectfont(#2)}%
  }%
}

\begin{tabular*}{\textwidth}{
@{\extracolsep{\fill}}
l
rrrr
c
@{\hspace{10pt}}
rr
c
@{}
}
\toprule

& \multicolumn{5}{c}{\textbf{LIBERO}}
& \multicolumn{3}{c}{\textbf{RoboTwin 2.0}} \\

\cmidrule(lr){2-6}
\cmidrule(lr){7-9}

\textbf{Model}
& \textbf{Spa.}
& \textbf{Obj.}
& \textbf{Goal}
& \textbf{Long}
& \textbf{Avg.}
& \textbf{Clean}
& \textbf{Random}
& \textbf{Avg.} \\

\midrule

\multicolumn{9}{@{}l}{\textit{Literature reports}} \\[1pt]

$\pi_0$ {\scriptsize\citep{yuan2026fastwam}}
& 96.8 & 98.8 & 95.8 & 85.2
& \avgval{94.1}
& 65.92 & 58.40
& \avgval{62.16} \\

$\pi_{0.5}$ {\scriptsize\citep{yuan2026fastwam}}
& 98.8 & 98.2 & 98.0 & 92.4
& \avgval{96.9}
& 82.74 & 76.76
& \avgval{79.75} \\

Motus {\scriptsize\citep{bi2025motus}}
& 96.8 & 99.8 & 96.6 & 97.6
& \avgval{97.7}
& 88.66 & 87.02
& \avgval{87.84} \\

LingBot-VA {\scriptsize\citep{li2026lingbotva}}
& 98.5 & 99.6 & 97.2 & 98.5
& \avgval{98.5}
& 92.93 & 91.55
& \avgval{92.24} \\

ImageWAM {\scriptsize\citep{zhang2026imagewam}}
& 97.2 & 99.2 & 98.8 & 98.4
& \avgval{98.4}
& 93.20 & 93.56
& \avgval{93.38} \\

\addlinespace[2pt]
\midrule

\multicolumn{9}{@{}l}{\textit{Our reproductions}} \\[1pt]

FastWAM
& 96.8
& 99.4
& 96.2
& 95.6
& \avgval{97.0}
& \textbf{91.44}
& 90.60
& \avgval{91.02} \\

\quad + JAC
& \textbf{98.0}
& \textbf{99.6}
& \textbf{98.2}
& \textbf{97.2}
& \jacavg{98.3}{+1.3}
& 91.32
& \textbf{91.16}
& \jacavg{91.24}{+0.22} \\

\addlinespace[3pt]

LeapBot-WA
& 96.2
& 99.6
& \textbf{97.6}
& 95.6
& \avgval{97.3}
& 90.92
& 92.06
& \avgval{91.49} \\

\quad + JAC
& \textbf{98.8}
& \textbf{99.8}
& 97.4
& \textbf{97.2}
& \jacavg{98.3}{+1.0}
& \textbf{91.46}
& \textbf{92.54}
& \jacavg{92.00}{+0.51} \\
\bottomrule
\end{tabular*}
\end{table*}

\paragraph{LIBERO: standard manipulation.}
Table~\ref{tab:main} reports the results on the four LIBERO suites. JAC raises the average success rate from 97.0 to 98.3 on FastWAM and from 97.3 to 98.3 on LeapBot-WA, corresponding to gains of 1.3 and 1.0 percentage points, respectively. Both hosts improve by 1.6 points on Long, while LeapBot-WA shows a slight 0.2-point decrease on Goal. Other gains vary by host: the largest increases occur on Goal for FastWAM (+2.0 points) and Spatial for LeapBot-WA (+2.6 points).

\paragraph{RoboTwin 2.0: bimanual manipulation.}
Our models use both clean and randomized demonstrations. When executing the first 16 predicted actions from each action chunk, JAC raises Avg. from 91.02 to 91.24 on FastWAM (\(+0.22\)) and from 91.49 to 92.00 on LeapBot-WA (\(+0.51\)). FastWAM gains 0.56 points on Random but loses 0.12 on Clean; LeapBot-WA gains 0.54 and 0.48 on Clean and Random. 

\paragraph{LIBERO-Plus: zero-shot robustness.}
Table~\ref{tab:plus} reports the seven perturbation categories. JAC improves all seven on both hosts, raising Total from 53.5 to 70.4 on FastWAM ($+16.9$) and from 73.1 to 77.3 on LeapBot-WA ($+4.2$). FastWAM's largest gains occur under Noise ($+34.1$) and Camera ($+24.5$), whereas LeapBot-WA improves most under Robot ($+5.7$), followed by Background and Noise ($+5.1$ each). The broader improvements under perturbation complement the smaller clean-LIBERO gains.

\begin{table}[!htb]
\centering
\footnotesize
\caption{\textbf{Zero-shot LIBERO-Plus.}
Success rates (\%) without Plus adaptation.
Aggregate is full-test Total for our rows and trial-level aggregation for
$\pi_0$/LAWA; $\dagger$ retains ImageWAM's reported Avg.;
equivalence to our episode-weighted Total is not assumed.
Parentheses denote changes in aggregate success rate (pp);
bold indicates gains.}
\label{tab:plus}

\vspace{2pt}

\setlength{\tabcolsep}{2.8pt}
\renewcommand{\arraystretch}{1.08}

\newcommand{\aggval}[1]{%
  \makebox[5.8em][c]{#1}%
}

\newcommand{\jacagg}[2]{%
  \makebox[5.8em][c]{%
    \textbf{#1}\,
    {\color{anchorgray}\fontsize{6.5}{7}\selectfont(#2)}%
  }%
}

\begin{tabular}{
@{}l
rrrrrrr
c
@{}
}
\toprule

\textbf{Model}
& \textbf{Cam.}
& \textbf{Robot}
& \textbf{Lang.}
& \textbf{Light}
& \textbf{Bg.}
& \textbf{Noise}
& \textbf{Layout}
& \textbf{Agg.} \\

\midrule

\multicolumn{9}{@{}l}{\textit{Literature reports}} \\[1pt]

$\pi_0$ {\scriptsize\citep{li2026lawa}}
& 13.8
& 6.0
& 58.8
& 85.0
& 81.4
& 79.0
& 68.9
& \aggval{53.6} \\

LAWA {\scriptsize\citep{li2026lawa}}
& 69.2
& 66.7
& 62.8
& 96.2
& 64.5
& 85.5
& 78.6
& \aggval{74.4} \\

\begin{tabular}[t]{@{}l@{}}
ImageWAM (FLUX.2 4B)\\[-1.5pt]
{\scriptsize\citep{zhang2026imagewam}}
\end{tabular}
& 80.8
& 50.3
& 91.4
& 98.1
& 85.5
& 93.8
& 80.5
& \aggval{83.1$^\dagger$} \\

\addlinespace[2pt]
\midrule

\multicolumn{9}{@{}l}{\textit{Our reproductions}} \\[1pt]

FastWAM
& 19.2
& 46.2
& 69.9
& 82.0
& 53.7
& 41.0
& 62.8
& \aggval{53.5} \\

\hspace{0.7em}+ JAC
& \textbf{43.7}
& \textbf{59.9}
& \textbf{82.5}
& \textbf{92.7}
& \textbf{62.7}
& \textbf{75.1}
& \textbf{76.3}
& \jacagg{70.4}{+16.9} \\

\addlinespace[3pt]

LeapBot-WA
& 33.8
& 75.1
& 87.8
& 93.4
& 90.1
& 54.7
& 76.7
& \aggval{73.1} \\

\hspace{0.7em}+ JAC
& \textbf{37.3}
& \textbf{80.8}
& \textbf{91.7}
& \textbf{95.9}
& \textbf{95.2}
& \textbf{59.8}
& \textbf{80.3}
& \jacagg{77.3}{+4.2} \\

\bottomrule
\end{tabular}

\end{table}

\subsection{Ablation Studies on LIBERO}
All three groups use FastWAM with a fixed $16R+16A$ JAC budget. The visibility study is conducted with $s=3$ and $\lambda_R=0.2$; the stride sweep fixes $\lambda_R=0.2$; and the measured weight sweep uses $s=2$, rather than the default stride-3 setting. Each group therefore examines one design factor while keeping the remaining settings fixed.

\paragraph{G1: cross-chain visibility.}
Directed \(R\rightarrow A\) access achieves the best performance at 98.3, compared with 98.2 for reverse access, 97.7 with no direct cross-chain access, and 97.5 for bidirectional access (Table~\ref{tab:visibility}). Among the tested visibility patterns, the directed \(R\rightarrow A\) design provides the strongest overall result, supporting an asymmetric information flow in which the execution chain can leverage the reference chain while the reference chain remains independent of the execution chain. We therefore adopt this directed reference-to-execution visibility as the default JAC design.

\begin{table}[!htb]
\centering
\small
\caption{G1: visibility on LIBERO at $s=3$, $\lambda_R=0.2$.
Permissions refer to direct attention before host intersection.
Padded targets are loss-masked but remain attention-visible.}
\label{tab:visibility}

\setlength{\tabcolsep}{3.5pt}
\renewcommand{\arraystretch}{1.08}

\begin{tabular}{
@{}l
@{\hspace{14pt}}
c
@{\hspace{12pt}}
c
@{\hspace{16pt}}
rrrrr
@{}
}
\toprule
\textbf{Information flow}
& \textbf{$A$ reads $R$}
& \textbf{$R$ reads $A$}
& \textbf{Spatial}
& \textbf{Object}
& \textbf{Goal}
& \textbf{Long}
& \textbf{Avg.} \\
\midrule

$R \leftrightarrow A$
& $\surd$
& $\surd$
& 98.0 & 99.4 & 97.0 & 95.4 & 97.5 \\

$R \rightarrow A$
& $\surd$
& $\times$
& 98.0 & 99.6 & 98.2 & 97.2 & \textbf{98.3} \\

No direct cross-chain
& $\times$
& $\times$
& 98.6 & 100.0 & 98.0 & 94.2 & 97.7 \\

$A \rightarrow R$
& $\times$
& $\surd$
& 98.4 & 99.6 & 98.0 & 96.6 & 98.2 \\

\bottomrule
\end{tabular}
\end{table}

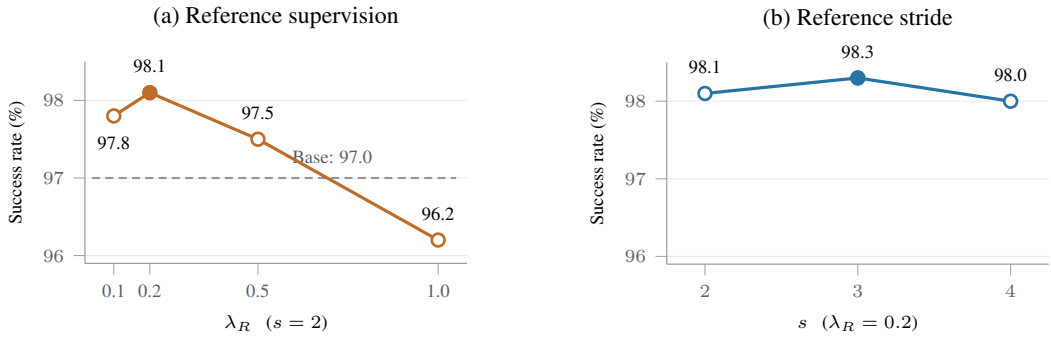
\begin{figure}[!htb]
\centering
\begin{tikzpicture}
\begin{axis}[width=0.475\textwidth,height=4.25cm,
axis lines=left,axis line style={draw=anchorgray!70,-},
tick align=outside,tick style={draw=anchorgray!60},
ymajorgrids=true,xmajorgrids=false,grid style={draw=gray!17},
tick label style={font=\scriptsize,text=anchorgray},label style={font=\scriptsize},
title style={font=\small,yshift=3pt},
scaled y ticks=false,yticklabel style={/pgf/number format/fixed,/pgf/number format/precision=1},
clip=false, ylabel={Success rate (\%)},title={(a) Reference supervision},xlabel={$\lambda_R$ \; ($s=2$)},xmin=0.02,xmax=1.08,xtick={0.1,0.2,0.5,1.0},xticklabels={0.1,0.2,0.5,1.0},ymin=95.9,ymax=98.5,ytick={96,97,98}]
\addplot[anchorgray!65,densely dashed,thick,no marks] coordinates {(0.04,97) (1.05,97)};
\node[font=\scriptsize,text=anchorgray,anchor=south west,yshift=2pt] at (axis cs:0.57,97){Base: 97.0};
\addplot[execorange,line width=1.2pt,mark=*,mark size=2.4pt,mark options={fill=white,line width=1pt}] coordinates {(0.1,97.8) (0.2,98.1) (0.5,97.5) (1,96.2)};
\addplot[only marks,execorange,mark=*,mark size=2.6pt] coordinates {(0.2,98.1)};
\node[font=\scriptsize,anchor=north,yshift=-4pt] at (axis cs:0.1,97.8){97.8};
\node[font=\scriptsize,anchor=south,yshift=4pt] at (axis cs:0.2,98.1){98.1};
\node[font=\scriptsize,anchor=south,yshift=4pt] at (axis cs:0.5,97.5){97.5};
\node[font=\scriptsize,anchor=south,yshift=4pt] at (axis cs:1,96.2){96.2};
\end{axis}
\end{tikzpicture}\hfill\begin{tikzpicture}
\begin{axis}[width=0.475\textwidth,height=4.25cm,
axis lines=left,axis line style={draw=anchorgray!70,-},
tick align=outside,tick style={draw=anchorgray!60},
ymajorgrids=true,xmajorgrids=false,grid style={draw=gray!17},
tick label style={font=\scriptsize,text=anchorgray},label style={font=\scriptsize},
title style={font=\small,yshift=3pt},
scaled y ticks=false,yticklabel style={/pgf/number format/fixed,/pgf/number format/precision=1},
clip=false, ylabel={Success rate (\%)},title={(b) Reference stride},xlabel={$s$ \; ($\lambda_R=0.2$)},xmin=1.75,xmax=4.25,xtick={2,3,4},ymin=95.9,ymax=98.5,ytick={96,97,98}]
\addplot[intentblue,line width=1.2pt,mark=*,mark size=2.4pt,mark options={fill=white,line width=1pt}] coordinates {(2,98.1) (3,98.3) (4,98.0)};
\addplot[only marks,intentblue,mark=*,mark size=2.6pt] coordinates {(3,98.3)};
\node[font=\scriptsize,anchor=south,yshift=4pt] at (axis cs:2,98.1){98.1};
\node[font=\scriptsize,anchor=south,yshift=4pt] at (axis cs:3,98.3){98.3};
\node[font=\scriptsize,anchor=south,yshift=4pt] at (axis cs:4,98.0){98.0};
\end{axis}
\end{tikzpicture}
\caption{\textbf{LIBERO sensitivity on FastWAM.} (a) Reference-supervision weight at $s=2$, with the no-JAC baseline shown as a separate dashed reference. (b) Reference stride at $\lambda_R=0.2$; the observed range is 0.3 pp. Measured settings share a restricted vertical scale; lines aid readability. Results use one training seed, without error bars.}
\label{fig:sensitivity}
\label{tab:weight}
\label{tab:stride}
\end{figure}

\paragraph{G2: reference-supervision weight.}
We vary the reference-supervision weight at \(s=2\). Setting \(\lambda_R\) to \(0.1, 0.2, 0.5,\) and \(1.0\) yields success rates of 97.8, 98.1, 97.5, and 96.2, respectively (Figure~\ref{fig:sensitivity}a). Performance peaks at the moderate setting \(\lambda_R=0.2\), while larger weights lead to progressively lower results, suggesting that the reference objective is most effective when used as a complementary supervision signal rather than dominating the action objective. The 97.0 FastWAM baseline is shown separately for reference and is not part of the weight sweep.

\paragraph{G3: reference stride.}
With $\lambda_R=0.2$, strides 2, 3, and 4 achieve 98.1, 98.3, and 98.0, respectively (Figure~\ref{fig:sensitivity}b), all outperforming the 97.0 FastWAM baseline. The best result is obtained at $s=3$, corresponding to a 1.3-point gain over the base model. The consistently strong performance across all three strides further shows that JAC is robust to the exact temporal spacing of the reference chain, while the intermediate stride provides the most favorable balance between temporal coverage and reference density.

\subsection{Computational Cost}
\label{sec:cost}
JAC reallocates the original $32A$ sequence into $16R+16A$ within the shared action expert, without adding trainable parameters or denoising steps. As shown in Table~\ref{tab:resources}, this restructuring introduces negligible inference overhead on both FastWAM and LeapBot-WA, with essentially unchanged latency and peak memory usage. These results indicate that JAC improves action organization while preserving the computational footprint of the original host models.

\begin{table}[!htb]
\centering
\caption{\textbf{Computational cost within each host.}
Model-call latency is wall-clock p50, including host-side scheduling and CPU--GPU transfers within inference. Measurements use the training hardware and configuration. Memory is peak allocated GPU memory.}
\label{tab:resources}

\vspace{2pt}
\fontsize{8.3}{10}\selectfont
\setlength{\tabcolsep}{4.5pt}
\renewcommand{\arraystretch}{1.08}

\begin{tabular*}{\textwidth}{@{\extracolsep{\fill}}llrrrr@{}}
\toprule
\textbf{Backbone} & \textbf{Configuration} & \textbf{Total params} & \textbf{Denoising} & \textbf{Latency} & \textbf{Peak memory}\\
& & \textbf{(M)} & \textbf{steps} & \textbf{(ms)} & \textbf{(GB)}\\
\midrule

FastWAM
& Base ($32A$)
& 11701.62
& 10
& 337.0
& 25.00 \\

& JAC ($16R+16A$)
& 11701.62
& 10
& 337.7
& 25.00 \\

\addlinespace[3pt]

LeapBot-WA
& Base ($32A$)
& 8428.11
& 10
& 438.6
& 26.13 \\

& JAC ($16R+16A$)
& 8428.11
& 10
& 435.1
& 26.13 \\

\bottomrule
\end{tabular*}
\end{table}

\subsection{Real-World Evaluation}
\begin{figure}[!htb]
\centering
\input{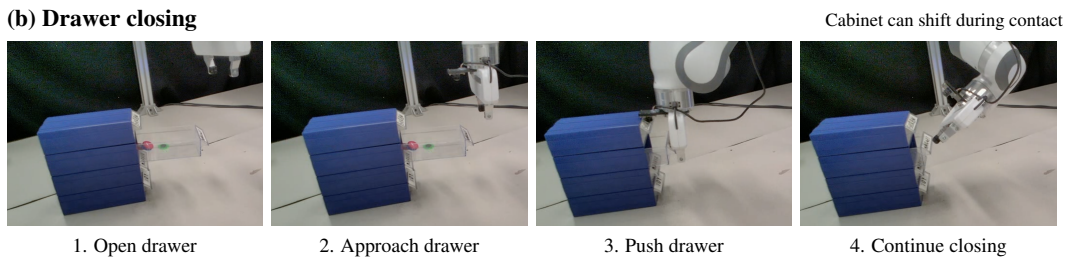}
\caption{\textbf{Real-world task sequences,} ordered left to right. (a) Apple-to-green-basket placement precedes green-block-to-blue-basket placement. (b) Drawer closing with possible cabinet displacement during contact. }
\label{fig:realworld}
\end{figure}

We evaluate two physical manipulation tasks (Figure~\ref{fig:realworld}). \textbf{T1: sequential sorting} requires placing an apple into the green basket, followed by placing a green block into the blue basket. Success requires both placements in the specified order. \textbf{T2: drawer closing} requires fully closing a drawer whose cabinet can shift during pushing, introducing motion of the manipulated fixture. 

Each model is evaluated for four rounds of 50 trials per task (200 trials per model per task), using one training seed; success rates are averaged across rounds. On sequential sorting / drawer closing, FastWAM achieves 82\% / 77\%, increasing to 93\% / 79\% with JAC; LeapBot-WA achieves 90\% / 85\%, increasing to 95\% / 90\%. The corresponding two-task averages, rounded to integer percentages, are 80\% / 86\% for FastWAM / FastWAM + JAC and 88\% / 93\% for LeapBot-WA / LeapBot-WA + JAC. These results show gains on both tasks and hosts.

The repeated 98.3 main/visibility/stride score denotes one shared configuration, not independent replication. LIBERO-only ablations do not isolate the source of LIBERO-Plus gains.

\section{Discussion and Limitations}

Reference stride and $\lambda_R$ jointly affect multiple training factors, limiting causal isolation in the current ablations. In addition, the sparse reference controls are not explicitly constrained to encode semantic intent, leaving their interpretation and broader transfer for future study.

\section{Conclusion}

We present \method{}, which reorganizes a fixed action-token budget into sparse reference and dense execution controls within a shared action expert. Across FastWAM and LeapBot-WA, \method{} improves performance on LIBERO, LIBERO-Plus, RoboTwin 2.0, and real-world tasks, with the largest gains under zero-shot perturbations. Ablations validate the main design choices, while matched inference shows negligible overhead. Overall, \method{} strengthens world-action models without additional model capacity or inference stages.
\clearpage
\subsection*{AI Use Statement}
In preparing this manuscript, generative AI tools assisted with drafting and English translation, literature discovery, clarification of method descriptions, feedback on experimental controls, figure editing, and LaTeX formatting. No empirical results were generated to replace experiments. The authors take responsibility for the final manuscript, including all AI-assisted text, figures, and claims.

\bibliographystyle{iclr2027_conference}
\bibliography{references}

@article{chi2023diffusion,
  title={Diffusion Policy: Visuomotor Policy Learning via Action Diffusion},
  author={Chi, Cheng and Xu, Zhenjia and Feng, Siyuan and Cousineau, Eric and Du, Yilun and Burchfiel, Benjamin and Tedrake, Russ and Song, Shuran},
  journal={arXiv preprint arXiv:2303.04137}, year={2023},
  url={https://arxiv.org/abs/2303.04137}
}

@article{zhao2023act,
  title={Learning Fine-Grained Bimanual Manipulation with Low-Cost Hardware},
  author={Zhao, Tony Z. and Kumar, Vikash and Levine, Sergey and Finn, Chelsea},
  journal={arXiv preprint arXiv:2304.13705}, year={2023},
  url={https://arxiv.org/abs/2304.13705}
}

@inproceedings{lipman2023flow,
  title={Flow Matching for Generative Modeling},
  author={Lipman, Yaron and Chen, Ricky T. Q. and Ben-Hamu, Heli and Nickel, Maximilian and Le, Matt},
  booktitle={International Conference on Learning Representations}, year={2023},
  url={https://arxiv.org/abs/2210.02747}
}

@article{yuan2026fastwam,
  title={{Fast-WAM}: Do World Action Models Need Test-time Future Imagination?},
  author={Yuan, Tianyuan and Dong, Zibin and Liu, Yicheng and Zhao, Hang},
  journal={arXiv preprint arXiv:2603.16666}, year={2026},
  url={https://arxiv.org/abs/2603.16666}
}

@article{assran2025vjepa2,
  title={{V-JEPA 2}: Self-Supervised Video Models Enable Understanding, Prediction and Planning},
  author={Assran, Mido and Bardes, Adrien and Fan, David and Garrido, Quentin and Howes, Russell and Komeili, Mojtaba and Muckley, Matthew and Rizvi, Ammar and Roberts, Claire and Sinha, Koustuv and Zholus, Artem and Arnaud, Sergio and Gejji, Abha and Martin, Ada and Hogan, Francois Robert and Dugas, Daniel and Bojanowski, Piotr and Khalidov, Vasil and Labatut, Patrick and Massa, Francisco and Szafraniec, Marc and Krishnakumar, Kapil and Li, Yong and Ma, Xiaodong and Chandar, Sarath and Meier, Franziska and LeCun, Yann and Rabbat, Michael and Ballas, Nicolas},
  journal={arXiv preprint arXiv:2506.09985}, year={2025},
  url={https://arxiv.org/abs/2506.09985}
}

@article{lynch2019play,
  title={Learning Latent Plans from Play},
  author={Lynch, Corey and Khansari, Mohi and Xiao, Ted and Kumar, Vikash and Tompson, Jonathan and Levine, Sergey and Sermanet, Pierre},
  journal={arXiv preprint arXiv:1903.01973}, year={2019},
  url={https://arxiv.org/abs/1903.01973}
}

@article{zawalski2024ecot,
  title={Robotic Control via Embodied Chain-of-Thought Reasoning},
  author={Zawalski, Micha{\l} and Chen, William and Pertsch, Karl and Mees, Oier and Finn, Chelsea and Levine, Sergey},
  journal={arXiv preprint arXiv:2407.08693}, year={2024},
  url={https://arxiv.org/abs/2407.08693}
}

@article{su2021roformer,
  title={{RoFormer}: Enhanced Transformer with Rotary Position Embedding},
  author={Su, Jianlin and Lu, Yu and Pan, Shengfeng and Murtadha, Ahmed and Wen, Bo and Liu, Yunfeng},
  journal={arXiv preprint arXiv:2104.09864}, year={2021},
  url={https://arxiv.org/abs/2104.09864}
}

@inproceedings{liu2023libero,
  title={{LIBERO}: Benchmarking Knowledge Transfer for Lifelong Robot Learning},
  author={Liu, Bo and Zhu, Yifeng and Gao, Chongkai and Feng, Yihao and Liu, Qiang and Zhu, Yuke and Stone, Peter},
  booktitle={Advances in Neural Information Processing Systems}, year={2023},
  url={https://arxiv.org/abs/2306.03310}
}

@inproceedings{fei2026liberoplus,
  title={{LIBERO-Plus}: A Progressive Robustness Benchmark for Visual-Language-Action Models},
  author={Fei, Senyu and Wang, Siyin and Shi, Junhao and Dai, Zihao and Cai, Jikun and Qian, Pengfang and Ji, Li and He, Xinzhe and Zhang, Shiduo and Fei, Zhaoye and Fu, Jinlan and Gong, Jingjing and Qiu, Xipeng},
  booktitle={Proceedings of the IEEE/CVF Conference on Computer Vision and Pattern Recognition}, year={2026}, pages={38574--38583},
  url={https://openaccess.thecvf.com/content/CVPR2026/papers/Fei_LIBERO-Plus_A_Progressive_Robustness_Benchmark_for_Visual-Language-Action_Models_CVPR_2026_paper.pdf}
}

@inproceedings{hao2025chd,
  title={{CHD: Coupled Hierarchical Diffusion for Long-Horizon Tasks}},
  author={Hao, Ce and Xiao, Anxing and Xue, Zhiwei and Soh, Harold},
  year={2025},
  url={https://proceedings.mlr.press/v305/hao25a.html},
  booktitle={Proceedings of The 9th Conference on Robot Learning},
  pages={1420--1451},
  volume={305}
}

@inproceedings{su2025densepolicy,
  title={{Dense Policy: Bidirectional Autoregressive Learning of Actions}},
  author={Su, Yue and Zhan, Xinyu and Fang, Hongjie and Xue, Han and Fang, Hao-Shu and Li, Yong-Lu and Lu, Cewu and Yang, Lixin},
  year={2025},
  url={https://openaccess.thecvf.com/content/ICCV2025/html/Su_Dense_Policy_Bidirectional_Autoregressive_Learning_of_Actions_ICCV_2025_paper.html},
  booktitle={Proceedings of the IEEE/CVF International Conference on Computer Vision},
  pages={14486--14495}
}

@inproceedings{chen2024diffusionforcing,
  title={{Diffusion Forcing: Next-token Prediction Meets Full-Sequence Diffusion}},
  author={Chen, Boyuan and Mart\'{i} Mons\'{o}, Diego and Du, Yilun and Simchowitz, Max and Tedrake, Russ and Sitzmann, Vincent},
  year={2024},
  url={https://arxiv.org/abs/2407.01392},
  booktitle={Advances in Neural Information Processing Systems},
  volume={37}
}

@article{zhou2024dinowm,
  title={{DINO-WM: World Models on Pre-trained Visual Features enable Zero-shot Planning}},
  author={Zhou, Gaoyue and Pan, Hengkai and LeCun, Yann and Pinto, Lerrel},
  year={2024},
  url={https://arxiv.org/abs/2411.04983},
  journal={arXiv preprint arXiv:2411.04983},
  eprint={2411.04983},
  archivePrefix={arXiv}
}

@inproceedings{du2023unipi,
  title={{Learning Universal Policies via Text-Guided Video Generation}},
  author={Du, Yilun and Yang, Mengjiao and Dai, Bo and Dai, Hanjun and Nachum, Ofir and Tenenbaum, Joshua B. and Schuurmans, Dale and Abbeel, Pieter},
  year={2023},
  url={https://arxiv.org/abs/2302.00111},
  booktitle={Advances in Neural Information Processing Systems},
  volume={36}
}

@inproceedings{ni2024cotdiffusion,
  title={{Generate Subgoal Images before Act: Unlocking the Chain-of-Thought Reasoning in Diffusion Model for Robot Manipulation with Multimodal Prompts}},
  author={Ni, Fei and Hao, Jianye and Wu, Shiguang and Kou, Longxin and Liu, Jiashun and Zheng, Yan and Wang, Bin and Zhuang, Yuzheng},
  year={2024},
  url={https://openaccess.thecvf.com/content/CVPR2024/html/Ni_Generate_Subgoal_Images_before_Act_Unlocking_the_Chain-of-Thought_Reasoning_in_CVPR_2024_paper.html},
  booktitle={Proceedings of the IEEE/CVF Conference on Computer Vision and Pattern Recognition},
  pages={13991--14000}
}

@article{liu2026leapbotwa,
 title={{LeapBot-WA}: World-Anchor Action Models via Predictive Latent Alignments},
 author={Liu, Pei and Zheng, Nan and Zhang, Lang and Peng, Daojie and Zhang, Yanan and Kong, Feilong and Feng, Mingyue and Liu, Jiachao and Wang, Yaonong and Chen, Qifeng and Ma, Jun},
 journal={arXiv preprint arXiv:2607.23969}, year={2026},
 eprint={2607.23969}, archivePrefix={arXiv},
 url={https://arxiv.org/abs/2607.23969}
}

@article{bi2025motus,
 journal={arXiv preprint arXiv:2512.13030},
 title={{Motus}: A Unified Latent Action World Model},
 author={Bi, Hongzhe and Tan, Hengkai and Xie, Shenghao and Wang, Zeyuan and Huang, Shuhe and Liu, Haitian and Zhao, Ruowen and Feng, Yao and Xiang, Chendong and Rong, Yinze and Zhao, Hongyan and Liu, Hanyu and Su, Zhizhong and Ma, Lei and Su, Hang and Zhu, Jun},
 year={2025}, eprint={2512.13030}, archivePrefix={arXiv}, url={https://arxiv.org/abs/2512.13030}}

@article{li2026lingbotva,
 journal={arXiv preprint arXiv:2601.21998},
 title={Causal World Modeling for Robot Control},
 author={Li, Lin and Zhang, Qihang and Luo, Yiming and Yang, Shuai and Wang, Ruilin and Han, Fei and Yu, Mingrui and Gao, Zelin and Xue, Nan and Zhu, Xing and Shen, Yujun and Xu, Yinghao},
 year={2026}, eprint={2601.21998}, archivePrefix={arXiv}, url={https://arxiv.org/abs/2601.21998}}

@article{chen2025robotwin2,
 journal={arXiv preprint arXiv:2506.18088},
 title={{RoboTwin 2.0}: A Scalable Data Generator and Benchmark with Strong Domain Randomization for Robust Bimanual Robotic Manipulation},
 author={Chen, Tianxing and Chen, Zanxin and Chen, Baijun and Cai, Zijian and Liu, Yibin and Li, Zixuan and Liang, Qiwei and Lin, Xianliang and Ge, Yiheng and Gu, Zhenyu and others},
 year={2025}, eprint={2506.18088}, archivePrefix={arXiv}, url={https://arxiv.org/abs/2506.18088}}

@article{zhang2026imagewam,
 journal={arXiv preprint arXiv:2606.19531},
 title={{ImageWAM}: Do World Action Models Really Need Video Generation, or Just Image Editing?},
 author={Zhang, Yuyang and Zhang, Wenyao and Qi, Zekun and Zhang, He and Lin, Haitao and Zhang, Jingbo and Mu, Yao and Yang, Xiaokang and Zeng, Wenjun and Jin, Xin},
 year={2026}, eprint={2606.19531}, archivePrefix={arXiv}, url={https://arxiv.org/abs/2606.19531}}

@article{li2026lawa,
 journal={arXiv preprint arXiv:2608.24882},
 title={Latent Action as Intention Enables Efficient Future Imagination for World Action Models},
 author={Li, Xiang and Zheng, Yupeng and Gu, Songen and Ma, Huailiang and Yu, Feng and Zheng, Yuhang and Nie, Xian and Yuan, Shanshuai and Zang, Yujie and Li, Weize and Tian, Shuai and Liu, Moyang and Zhang, Ya-Qin and Ding, Wenchao},
 year={2026}, eprint={2608.24882}, archivePrefix={arXiv}, url={https://arxiv.org/abs/2608.24882v2}}

@article{yu2026maskwam,
 title={{MaskWAM}: Unifying Mask Prompting and Prediction for World-Action Models},
 author={Yu, Hanyang and Lin, Haitao and Zhang, Jingbo and Zhang, Wenyao and Gu, Chenghao and Li, Heng and Tan, Ping},
 journal={arXiv preprint arXiv:2606.13515}, year={2026}, url={https://arxiv.org/abs/2606.13515}
}

@article{yan2026robustwam,
 title={{Robust-WAM}: Bridging Generative Pretraining and Semantic Foresight in World-Action Models},
 author={Yan, Haodong and Li, Junfeng and He, Junjie and Zhong, Zhide and Yu, MingMing and Song, Wenxuan and Zhu, Jiaguan and Zheng, Yangyang and Du, Yuqiao and You, Jiadi and Cai, Yingjie and Yan, Xu and Zhao, Guanyi and Liu, Bingbing and Li, Haoang},
 journal={arXiv preprint arXiv:2608.05903}, year={2026}, url={https://arxiv.org/abs/2608.05903}
}

@article{yang2026memorywam,
 title={{MemoryWAM}: Efficient World Action Modeling with Persistent Memory},
 author={Yang, Sizhe and Mu, Juncheng and Wei, Tianming and Lu, Chenhao and Li, Xiaofan and Xu, Linning and Xue, Zhengrong and Yuan, Zhecheng and Lin, Dahua and Pang, Jiangmiao and Xu, Huazhe},
 journal={arXiv preprint arXiv:2606.20562}, year={2026}, url={https://arxiv.org/abs/2606.20562}
}
\clearpage
\appendix
\section{Experimental Details and Reporting Scope}
\label{app:controls}
\subsection{Configurations and Target Spans}
All reported JAC main configurations use $n_R=n_A=16$, $s=3$, and $\lambda_R=0.2$. The reference indices are $\{0,3,\ldots,45\}$ and execution indices are $\{0,\ldots,15\}$. For a fully available target window, six times overlap, giving 26 distinct controls across 32 tokens. The comparison protocol executes $K=10$ controls on LIBERO/LIBERO-Plus and $K=16$ on RoboTwin, with the same replanning interval for each base/JAC pair. The method and Figure~\ref{fig:timemask} use this same default. The measured weight ablation at $s=2$ is an explicitly identified exception.

\begin{table}[h]
\centering\small
\caption{Reported JAC configuration groups. All use FastWAM unless another host is listed. All results use one training seed.}
\label{tab:reuse}
\begin{tabular}{llll}
\toprule Evaluation & Host & Stride & Reference weight\\\midrule
Main performance & FastWAM; LeapBot-WA &3&0.2\\
G1 visibility & FastWAM &3&0.2\\
G2 weight & FastWAM &2&0.1, 0.2, 0.5, 1\\
G3 stride & FastWAM &2, 3, 4&0.2\\
\bottomrule
\end{tabular}
\end{table}

Each host is reproduced using the training hyperparameters in its original paper~\citep{yuan2026fastwam,liu2026leapbotwa}, with the JAC target, position, visibility, and loss changes specified in Section 3. All local results use one common training seed. Training and evaluation run on a system with eight NVIDIA H100 GPUs. RoboTwin uses both clean and randomized demonstrations; the four local models are re-evaluated with replan$=16$. Literature rows retain their published protocols and are not treated as matched-budget experiments.

\subsection{Reference-Stride Eligibility}
\label{app:stride}
For $I_R=\{0,s,\ldots,15s\}$, strides 2, 3, and 4 require minimum future-label spans of 31, 46, and 61 controls. The confirmed loader window for the default $s=3$ is 48. With fully available labels, their overlaps with the execution interval are 8, 6, and 4, giving 24, 26, and 28 distinct target times. These are target-layout counts, not guaranteed counts of real labels at trajectory boundaries.

All stride conditions use identical trajectory starts. When a requested future control is unavailable, the sequence is padded; no common 61-valid-control filter is applied to select a different cohort for this comparison. Thus the starting-state distribution is held fixed, while reference span, target overlap, and the frequency of boundary padding can change together. The confirmed fill and masking rules are described in Appendix~\ref{app:padding}. Padding frequency was not quantified by stride. The non-monotonic stride results therefore should not be attributed solely to coverage. They also do not establish superiority over dense repeated near-term targets.

\subsection{Experimental Settings}
\label{app:parameters}
Host-specific optimization, training schedules, and baseline hyperparameters follow FastWAM and LeapBot-WA~\citep{yuan2026fastwam,liu2026leapbotwa}. Table~\ref{tab:parameters} records the shared infrastructure and JAC-specific settings. Single-seed results do not characterize training-seed variability.
\begin{table}[htbp]
\centering\small
\caption{Implementation and evaluation settings. Host-specific hyperparameters follow the cited original papers; execution settings below apply to our paired comparisons.}
\label{tab:parameters}
\begin{tabular}{p{0.46\linewidth}p{0.45\linewidth}}
\toprule Parameter & Setting\\\midrule
Host training hyperparameters & Original FastWAM / LeapBot-WA recipes\\
Training and evaluation hardware & Eight NVIDIA H100 GPUs\\
Training seeds & One common seed across local experiments\\
Action-token budget & Base: $32A$; JAC: $16R+16A$\\
Default reference settings & $s=3$, $\lambda_R=0.2$, $R\rightarrow A$\\
Weight sweep & $s=2$; $\lambda_R\in\{0.1,0.2,0.5,1\}$\\
LIBERO / LIBERO-Plus replan & 10 executed actions\\
RoboTwin replan & 16 executed actions for all local models\\
RoboTwin training data & Clean and randomized demonstrations\\
LIBERO-Plus evaluation & Zero-shot; full-test episode aggregation\\
Default loaded action window & 48 controls; reference indices $0,3,\ldots,45$\\
World branch across $s=2,3,4$ & Fixed baseline frame indices, counts, steps, loss weights\\
Real-world evaluation & 4 rounds $\times$ 50 trials per model per task\\
\bottomrule
\end{tabular}
\end{table}

\subsection{Padding Implementation}
\label{app:padding}
For each requested future index, the loader clamps an out-of-range index to the final legal action in that demonstration and records the position in \texttt{action\_is\_pad}. For LIBERO, \texttt{delta\_action\_dim\_mask} sets the first six components of padded controls to zero; the non-delta gripper component retains its final legal value. The loader then applies the normal action transformations, dataset normalization, and action-dimension merging to both valid and padded controls. Padding thus has an ordinary target tensor in normalized action space, and its delta components need not be numerically zero there.

The default full window contains 48 controls. Reference targets and masks select indices $0,3,\ldots,45$; execution targets and masks select $0,1,\ldots,15$. Padding slots undergo the same flow noising and network forward pass, but the factors $1-P_{L,k}$ in Eq.~\eqref{eq:chainloss} remove their direct target errors from the loss. Each chain loss averages over action dimensions and its valid tokens, using a denominator clamped to at least one token before applying the shared timestep weight. If a chain has no valid targets, its direct loss is zero.

The padding mask is not incorporated into the Action DiT self-attention key-padding mask. Padded references can serve as keys and values to permitted reference and execution queries. They can consequently influence valid predictions and receive indirect gradients through those predictions. This fixed-length treatment must be distinguished from deleting padded tokens or excluding them from attention.

\subsection{Loss Accounting}
\label{app:losscontrol}
For each sample and chain, the implementation first averages elementwise squared flow error over the $d$ action dimensions, then sums over non-padding tokens and divides by $\max(N_L,1)$. It multiplies the result by the common timestep weight $w(\tau)$ and finally averages over the batch. This reduction yields Eq.~\eqref{eq:chainloss}; it is not an unnormalized masked sum.

The inherited scheduler uses a Gaussian-shaped weight centered at the midpoint of its training timestep range:
\begin{equation}
w(\tau)=\frac{\exp\!\left[-2\left((\tau-N/2)/N\right)^2\right]-y_{\min}}{c},\qquad N=1000,
\end{equation}
where $y_{\min}$ and $c$ are the scheduler's offset and normalization constants, giving approximately unit mean under its training sampling distribution. Both chains use the same sampled timestep and scheduler-mapped noise level. This weighting is inherited from the host, rather than introduced by JAC; its approximately unit mean does not imply an unchanged expected weighted loss, since residuals depend on timestep.

With execution coefficient 1 and fixed world-loss coefficients, $\lambda_R$ weights the normalized reference objective relative to the execution objective. Valid-token averaging removes the direct multiplicative effect of valid-token count, but does not guarantee equal loss magnitudes or gradient norms across chains. Padding can alter which physical times receive supervision and can affect predictions through attention-visible tokens. Changing $\lambda_R$ also changes the action objective relative to the world objective; separate role-weighted means are not generally equivalent to a uniformly weighted single-chain mean.

No loss-reweighted single-chain control or zero-weight JAC result is reported. Consequently the base/JAC gain reflects the combined effects of target allocation, visibility, positions, and loss weighting. The weight sweep supports sensitivity to the tested positive coefficients; it does not establish the necessity of direct reference supervision. The no-JAC base remains a separate comparator rather than an assumed zero-weight endpoint.

\subsection{Metrics and Run Identity}
\label{app:reconcile}
\label{app:measurements}
LIBERO Avg. is the equally weighted mean of Spatial, Object, Goal, and Long, rounded to one decimal place. LIBERO-Plus Total aggregates the full test evaluation and is not reconstructed by averaging the seven displayed categories. All local results use a single common training seed. LIBERO-Plus category counts need not be equal, so its Total is not a macro-average of the displayed categories. Reported differences are descriptive, not significance estimates. Real-world rates average four equal-size evaluation rounds of 50 trials per model per task; these rounds do not constitute independent training seeds.

The main FastWAM + JAC aggregate at $s=3$, $\lambda_R=0.2$ is 98.3/70.4, consistent with the corresponding visibility and stride rows. These appearances do not constitute independent replications. The $s=2$, $\lambda_R=0.2$ weight/stride result remains 98.1/63.0. No conclusions about seed variability follow from the displayed decimal precision. Neither inference latency nor peak-memory improvements are claimed from fixed token counts.

\subsection{RoboTwin Evaluation Protocol}
All four local models use clean and randomized training demonstrations and are evaluated with replan$=16$. FastWAM achieves 91.44/90.60/91.02 on Clean/Random/Avg., versus 91.32/91.16/91.24 with JAC. LeapBot-WA achieves 90.92/92.06/91.49, versus 91.46/92.54/92.00 with JAC. Table~\ref{tab:main} reports these completed paired comparisons.

\subsection{Host Integration}
\label{app:integration}
JAC constructs reference/execution targets, concatenates their noisy states, maps physical-time positions, applies a role mask intersected with host legality, and separates the resulting outputs within the existing action expert. The local current-semantic-cache implementation freezes the V-JEPA encoder and its adapter while training the action expert, proprioception encoder, and JEPA context projector used for semantic anchoring. The host-specific training configuration follows its original recipe; this semantic-cache description is not assumed to specify every host.

In both evaluated hosts, video-frame indices, frame counts, future world-prediction steps, and world-loss coefficients are identical between base and JAC and remain fixed across reference strides $s=2,3,4$. The data pipeline retains each host's baseline video-sampling grid when the future-action window is extended. Thus a longer reference span does not introduce additional world targets.

Figure~\ref{fig:timemask} depicts an action-independent semantic context. In hosts with feedback through world tokens or caches, direct role masking does not guarantee global independence. Table~\ref{tab:resources} measures model-call latency, including host-side scheduling and CPU--GPU transfers within inference, rather than a complete robot observation-to-command cycle. Action-token counts alone do not determine runtime.

\subsection{Forward-Dependency Analysis}
\label{app:dependency}
Suppose that all context visible to $R$ within a query is independent of the execution-chain initialization, and no token mixing, cache update, or solver operation introduces a reverse path. At fixed parameters and conditions, induction over layers and solver updates gives
\begin{equation}
\frac{\partial R^{(q)}}{\partial A^{\rm init}}=0,
\end{equation}
where $q$ indexes solver updates. A fixed, action-independent semantic cache satisfies these conditions. This is a conditional statement about forward computation, not statistical independence: the chains share conditions and parameters, and execution losses can backpropagate through reference representations. It also does not exclude effects of past executed actions on later observations. In coupled hosts, an $A\rightarrow W\rightarrow R$ path may violate the stated condition; removing that feedback or copying context would change the host and require separate cost reporting.

\subsection{Sampling Procedure}
\label{app:sampling}
The two chains are updated jointly within the existing action expert; only the execution prefix is sent to the environment. The host schedule determines any world-state and cache updates.

\begin{algorithm}[!htb]
\caption{JAC sampling with the host process retained}
\label{alg:jac}
\begin{algorithmic}[1]
\Require Conditions $c_t$, host schedule, indices $I_R,I_A$
\State Initialize world states $W$ and history $C$ as in host
\State Independently sample $R,A\sim\mathcal N(0,I)$
\For{each update specified by the host}
\State Map physical-time positions; compute Eq.~\eqref{eq:mask}
\State Replace action inputs with $[R;A]$ in host modules
\State Jointly predict and update $R,A$ on action updates
\State Update world states and caches on the host schedule
\EndFor
\State Inverse-transform $A$; execute its first $K$ controls
\end{algorithmic}
\end{algorithm}

\section{Limits of Mechanistic Interpretation}
\label{app:evidence}
The three ablations assess direct visibility, positive reference-loss weights, and temporal strides. They support configuration sensitivity, but do not identify reference semantics or prove abstract planning. Sparse controls may act as auxiliary predictive targets, and blocked direct access may retain indirect communication. Isolating target duplication, reference content, supervision allocation, and optimization effects requires additional matched controls; no unperformed mechanism study is claimed as evidence.

\end{document}